\documentclass[sigplan,screen]{acmart}
\AtBeginDocument{%
  \providecommand\BibTeX{{%
    \normalfont B\kern-0.5em{\scshape i\kern-0.25em b}\kern-0.8em\TeX}}}

\usepackage{subcaption}
\usepackage{listings}
\usepackage{comment}
\usepackage{url}
\usepackage{amsmath,amsfonts}%,amssymb
\usepackage{bbm}
\usepackage[noend]{algpseudocode}
\usepackage{algorithm}
\usepackage{tcolorbox}
\usepackage{tikz}
\usepackage{float}
\usepackage[mode=buildnew]{standalone}
\usetikzlibrary{matrix,chains,positioning,decorations.pathreplacing,arrows}
\usepackage{pgfplots}
\pgfplotsset{compat=1.10}
\usetikzlibrary{shapes.geometric,arrows,fit,matrix,positioning, shapes, calc}
\usepackage{ulem}
\usepackage{multirow}
\pgfarrowsdeclarecombine{twotriang}{twotriang}%    double headed arrow
{stealth'}{stealth'}{stealth'}{stealth'}
\newcommand{\Ignore}[1]{}

\begin{document}

\title{QNNVerifier: A Tool for Verifying Neural Networks using SMT-Based Model Checking}

\author{Xidan Song}
% \email{xidan.song@postgrad.manchester.ac.uk}
\author{Edoardo Manino}
% \email{edoardo.manino@manchester.ac.uk}
\affiliation{%
  \institution{University of Manchester}
  \streetaddress{Kilburn Building, Oxford Road}
  \city{Manchester}
  \country{UK}
  \postcode{ M13 9PL}
}

\author{Luiz Sena, Erickson Alves, \\ Eddie de Lima Filho, Iury Bessa}
% \email{ {coelho.luiz.sena, erickson.higor }@gmail.com}
% \author{Erickson Alves}
% \email{erickson.higor@gmail.com}
% \author{Eddie de Lima Filho, Iury Bessa}
% \email{iurybessa@ufam.edu.br}
% \author{Eddie de Lima Filho}
% \email{eddie_batista@yahoo.com.br}
\affiliation{%
  \institution{ University of Amazonas}
  \streetaddress{Av. Gal. Rodrigo Oct\'avio, 1200}
  \city{Manaus}
  \country{Brazil}
  \postcode{69067-005}
}

\author{Mikel Luj\'{a}n, Lucas Cordeiro}
\email{lucas.cordeiro@manchester.ac.uk}
\affiliation{%
  \institution{University of Manchester}
  \streetaddress{Kilburn Building, Oxford Road}
  \city{Manchester}
  \country{UK}
  \postcode{ M13 9PL}
}
\renewcommand{\shortauthors}{Song et al.}

\begin{abstract}
 \textit{QNNVerifier} is the first open-source tool for verifying implementations of neural networks that takes into account the finite word-length (i.e.\@ quantization) of their operands. The novel support for quantization is achieved by employing state-of-the-art software model checking (SMC) techniques. It translates the implementation of neural networks to a decidable fragment of first-order logic based on satisfiability modulo theories (SMT). The effects of fixed- and floating-point operations are represented through direct implementations given a hardware-determined precision. Furthermore, \textit{QNNVerifier} allows to specify bespoke safety properties and verify the resulting model with different verification strategies (incremental and \textit{k}-induction) and SMT solvers. Finally, \textit{QNNVerifier} is the first tool that combines invariant inference via interval analysis and discretization of non-linear activation functions to speed up the verification of neural networks by orders of magnitude. A video presentation of \textit{QNNVerifier} is available at \url{https://youtu.be/7jMgOL41zTY}.
\end{abstract}





\maketitle

%-------------------------------------------------------
\section{Introduction}
\label{sec:intro}
%-------------------------------------------------------

% Motivation

Artificial Neural Networks (ANNs) are machine learning models that can solve an extensive range of problems, including pattern recognition, decision making, and approximation of physical systems~\cite{bishop2006PRML}. Unfortunately, ANNs may be fragile to adversarial input perturbations, which induces erratic outputs~\cite{huang2017safety}. Moreover, the black-box nature of ANNs makes them challenging to interpret and debug~\cite{LundbergL17}. This further compounds the risk of unwanted behaviors remaining undetected. As the deployment of ANNs in safety-critical applications becomes widespread, there has been a growing interest in verification methods to certify their behavior.

% Background

The existing approaches for ANNs verification can be divided into three different families. First, \textit{optimization-based} approaches handle the required safety properties as constraints and search for the most damaging adversarial input that satisfies them~\cite{Tjeng2018}. However, ANNs present challenging non-linear and non-convex optimization problems. Some approaches solve a relaxation of the original problem to scale but sacrifice the completeness of their results~\cite{Dvijotham2018, Fazlyab2020}.

Second, \textit{reachability-based} approaches take a safe set of inputs and propagate it through the layers of an ANN. Safety violations are found when the output set exceeds the boundaries of a given safe region~\cite{Xiang2018}. However, computing the exact output sets is challenging. Thus, existing approaches over-approximate the result using symbolic~\cite{Wang2018} or set-theoretic techniques~\cite{Xiang2018,Tran2020}, which are sound but not complete.

Last, \textit{satisfiability-based} approaches encode both the ANN and the desired safety property into a single logic formula and then check whether a counterexample exists. In this regard, only binarized neural networks~\cite{Narodytska2018} can be directly encoded into propositional logic and verified with SAT solvers. For general ANNs, Satisfiability Modulo Theories (SMT) offer a more compact encoding as they can describe the semantics of ANNs (real-valued arithmetic, ReLU activations, etc.) using a decidable fragment of first-order logic~\cite{katz2017reluplex,amir2021smt}.

% Challenges

While satisfiability-based approaches are exact, the resulting verification problem is challenging to solve~\cite{pulina2012challenging}. Moving from the idealized mathematical model of ANNs (infinite precision) to their quantized implementation (floating- or fixed-point) makes the computational problem even harder~\cite{Henzinger2020}. Given the interest in deploying extremely quantized ANNs for low-power applications~\cite{Guo2018}, a few recent studies have proposed new fixed-point SMT background theories~\cite{Giacobbe2020,Baranowski2020}. Unfortunately, these studies do not publish their source code and respective tool, thus limiting the impact of their approaches on real-world verification of ANNs.

% Contribution

Given this, we present the first open-source tool, known as \textit{QNNVerifier}, which efficiently verifies the quantized implementation of ANNs both in fixed- and floating-point.  \textit{QNNVerifier} checks the source code of the ANN rather than its abstract mathematical model. By doing so, we can verify the actual ANN implementation and leverage several recent advances in software verification that dramatically speed up the verification process. More in detail, \textit{QNNVerifier} has the following unique features:

% Features

\begin{itemize}
    \item The input ANN can be in various formats, including ONNX, NNET, Keras, Tensorflow, or native C code.
    \item The input ANN can include any type of layer architecture supported by \texttt{keras2c}.\footnote{\url{https://github.com/f0uriest/keras2c}}
    \item The input ANN can include any non-linear activation function, as our tool replaces them with look-up tables to speed up the verification process.
    \item The user can specify any type of safety property expressible in plain C code. Here, we give examples of robustness properties.
    \item Our tool supports both IEEE 754 floating-point and any fixed-point precision, as long as the whole network uses the same precision.
    \item In the backend, our tool uses the state-of-the-art software verification tool ESBMC
    % , which supports several SMT solvers
    ~\cite{GadelhaMMC0N18}. Thus, any future improvement in ESBMC and its SMT solvers will automatically translate into benefits for \textit{QNNVerifier}.
\end{itemize}

% originality

The code of \textit{QNNVerifier} is available at \url{https://zenodo.org/record/5724254#.YZ47YdDP13g}.
.
%-------------------------------------------------------
\section{Tool description}
\label{sec:tool}
%-------------------------------------------------------

%-------------------------------------------------------
\subsection{Technical approach}
\label{subsec:tech}
%-------------------------------------------------------

% discretization, C OM, quantization, SMT/MC

\paragraph{C code abstract model for ANNs}
ANNs are highly parallel models built by combining basic building blocks called neurons. The output $y_{k} \in \mathbbm{R}$ of each neuron $k$ is a function of its inputs $x_k\in \mathbbm{R}^m$ defined by the composition of two functions $u_k$ and $\mathcal{N}_k$. The activation potential $u_k$ is an affine projection of the $m$ local inputs $x_k$. The activation function (AF) $\mathcal{N}_k$ is a non-linear mapping of the potential $u_k$:
\begin{equation}
% \small
\label{eq:anncalc_aff}
    y_k = \mathcal{N}_k \circ  u_k (x_k), \;
    u_k\left(x_k\right) =  {\textstyle \sum_{j=1}^{m}{w_{j,k} x_{j,k}} + b_k}.
\end{equation}

In this paper, we use the C language as an abstract model of the ANN implementation. It allows us to model each operation in its quantized form explicitly (fixed- or floating-point) and apply mature software verification techniques.

\paragraph{Fixed-point operational model for ANNs}

For low-power applications, it is beneficial to implement ANNs in fixed-point arithmetic, rather than the more energy-hungry floating-point format~\cite{Guo2018}. To cover this case, we allow the user to specify which operations in the ANN implementation should be verified in fixed-point format. Furthermore, we let the user customize the fixed-point representation by freely choosing the number of integer and fractional bits. Our approach is based on the fixed-point models in~\cite{Chaves2019}: all arithmetic operations ($+$, $-$, $*$, and $/$) are modeled as an SMT background theory, thus making them compatible with SMT solvers.

As an example, Figure~\ref{fig:fxp-sample-conversion-code-fixed} shows a code snippet that computes the activation potential of a single neuron in fixed-point. In particular, \texttt{fxp\_float\_to\_fxp} transforms a type \texttt{float} into a  type \texttt{fxp\_t} (fixed point), and both \texttt{fxp\_add} and \texttt{fxp\_mult} perform additions and multiplications in fixed-point. The integer and fractional precisions are specified globally as a preprocessor directive (not shown here).

% \resizebox{0.9\columnwidth}{!}{
\begin{figure}[ht]
% \begin{subfigure}{\columnwidth}
%\lstset{morekeywords={fxp\_t,fxp\_float\_to\_fxp,fxp\_add,fxp\_mult,if}}
\begin{lstlisting}[numbers=left,xleftmargin=2em,frame=single,framexleftmargin=1.5em, basicstyle=\tiny]  
fxp_t potential(float *w, float *x, unsigned int w_len,
                unsigned int w_len, float b) {
  if (w_len != x_len) {
    return 0;
  }
  fxp_t result = 0;
  for (unsigned int i = 0; i < w_len; ++i) {
    fxp_t w_fxp = fxp_float_to_fxp(w[i]);
    fxp_t x_fxp = fxp_float_to_fxp(x[i]);
    result = fxp_add(result,fxp_mult(w_fxp,x_fxp));
  }
  fxp_t b_fxp = fxp_float_to_fxp(b);
  result = fxp_add(result, b_fxp);
  return result;
  }
\end{lstlisting}
% \caption{}%Fixed-point version}
% \end{subfigure}
\caption{C-code for fixed-point computation of $u_{k}$. \label{fig:fxp-sample-conversion-code-fixed}}
% \label{fig:fxp-sample-conversion-code}
\end{figure}
% }

\paragraph{Discretization of AFs}

Among all non-linear AFs, only the ReLU function can be efficiently encoded in SMT, as it requires a single if-then-else operation. To support other non-linear AFs, we convert them into look-up tables, which dramatically speeds up the verification time. To this end, assume that the AF $\mathcal{N}:\mathcal{U} \mapsto \mathbb{R}$ is piecewise Lipschitz continuous, thus there are $a$ locally Lipschitz continuous functions $\mathcal{N}_{i}:\mathcal{U}_{i} \mapsto \mathbb{R}$ for $i\in \mathbb{N}_{\leq a}$ with disjoint intervals $\mathcal{U}_{i}\subset \mathbb{R}$ and Lipschitz constant $\lambda_{i}$.  First, we discretize each $\mathcal{U}_{i}$ with a countable set $\tilde{\mathcal{U}}_{i} \subset \mathcal{U}_{i}$. Then, we build a lookup table for rounding $\mathcal{N}_{i}(u)$ to $\tilde{\mathcal{N}}_{i}(u):\mathcal{U}_{i}\mapsto \mathcal{R}$, thus rounding $\mathcal{N}(u)$ to  $\tilde{\mathcal{N}}(u)\in \left\lbrace \tilde{\mathcal{N}}_{1}(u), \ldots ,\tilde{\mathcal{N}}_{a}(u)  \right\rbrace$. The lookup table contains uniformly distributed $N_{i}$ samples within $\mathcal{U}_{i}$ to guarantee $\| \tilde{\mathcal{N}}_{i}(u) -  \mathcal{N}_{i}(u)\| \leq \epsilon $ for a given $\epsilon$. 

\begin{figure*}[t]
    \centering
    \includegraphics[width=\textwidth]{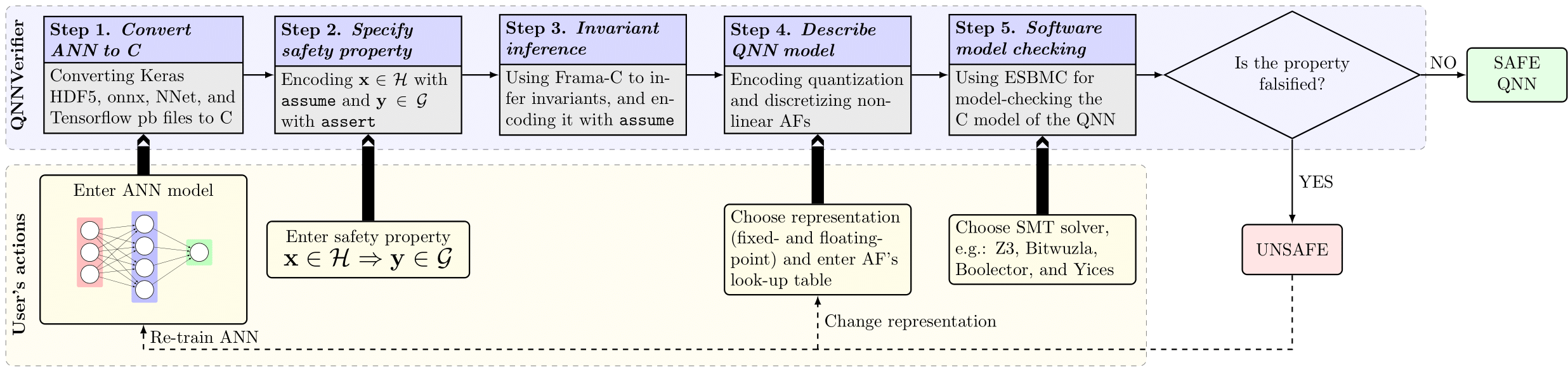}
    \caption{QNNVerifier architecture and usage.} \label{fig:qnnverifier}
\end{figure*}

\paragraph{Safety properties}
The following steps are used for specifying safety properties. First, the concrete ANN input is replaced by a non-deterministic one, which is enabled by the command \texttt{nondet\_float()} from the ESBMC~\cite{GadelhaMMCN20} to prepare \textit{QNNVerifier} to expect any possible value: 
\begin{align}
    & \texttt{float x\_1 = nondet\_float()}\\
    & \texttt{float x\_2 = nondet\_float()} \nonumber
\end{align}
\noindent Then, let us consider safety properties in the implication form $\mathbf{x}\in\mathcal{H}\implies \mathbf{y}\in\mathcal{G}$, e.g.\@ robustness properties. We model the premise $\mathbf{x}\in\mathcal{H}$ with the instruction \texttt{assume} by specifying the input domain $\mathcal{H}$. For example, the input domain defined by $x_1\in[0,2]$ and $x_2\in[-\frac{1}{2},+\frac{1}{2})$ is encoded as:
\begin{align}
    & \label{eq:safety_prop_input_assume}
    \texttt{assume(x\_1\;>=\;0\;\&\&\;x\_1\;<=\;2)}, \\
    & \texttt{assume(x\_2\;>=\;-0.5\;\&\&\;x\_2\;<\;0.5)}.  \nonumber
\end{align}

\noindent Similarly, we model the consequent with the instruction \texttt{assert} by specifying the safe domain $\mathcal{G}$ for the output $\mathbf{y}$. For instance, consider the binary classifier with outputs $y_1$ and $y_2$ denoting the score of each class. We can require that all outputs are assigned to class $2$ as follows:
\begin{equation}
    \texttt{assert(y\_2\;>\;y\_1)}.
\end{equation}
\noindent By combining different input and output domains, the user can specify a large variety of safety properties. If that does not suffice, the instructions \texttt{assume} and \texttt{assert} can be freely used to express any custom property in plain C code.

\paragraph{Invariant inference}

Once a safety property is specified, we prune the search space by inserting additional \texttt{assume} instructions (see Figure \ref{fig:insert_c_invariant}). More specifically, we use invariant analysis methods, which take the input domain $\mathcal{H}$ and propagate it through the ANN. In this way, we compute upper and lower bounds on the value of each neuron (intervals), which are guaranteed to hold during the verification process. To do so, we employ the open-source tool FRAMA-C with the evolved value analysis (EVA) plugin~\cite{blanchard2018}.

\begin{figure}[h]
\begin{lstlisting}[numbers=left,xleftmargin=2em,frame=single,framexleftmargin=1.5em]
float layer1[50];
layer1[0]= (5.40062e-02)*x[0] + (-2.61092e+00)*x[1] + (-1.80027e-01)*x[2] + (2.42194e-01)*x[3] + (1.41407e-01)*x[4] + (2.27630e-01);
if (layer1[0] < 0) layer1[0] = 0; // ReLU
__ESBMC_assume((layer1[0] >= -0.0) && (layer1[0] <= 0.256393373013));
layer1[1]= (-1.12374e+00)*x[0] + (2.63619e-02)*x[1] + (-9.17929e-03)*x[2] + (5.56230e-02)*x[3] + (-3.27635e-01)*x[4] + (-1.88762e-01);
if (layer1[1] < 0) layer1[1] = 0; // ReLU
...
\end{lstlisting}
\caption{Example of invariant in the ANN code (Line 4). \label{fig:insert_c_invariant}}
\end{figure}

%-------------------------------------------------------
\subsection{Verification framework}
\label{subsec:overview}
%-------------------------------------------------------

The overall architecture of \textit{QNNVerifier} is illustrated in Fig.~\ref{fig:qnnverifier}. We describe it by following the typical workflow of our users.

In step 1, \textit{QNNVerifier} converts a given ANN model into C code. The user can invoke our tool by graphical interface or command line under the guidance of the prompt message. Currently, supported formats include Keras HDF5, ONNX, TensorFlow PB, and NNET. Note that the latter can only represent fully-connected feedforward neural networks.

In step 2, the user describes the safety properties by inserting \texttt{assume} statements to constrain the search-space exploration and \texttt{assert} statements to specify the safety property in the C code.
In step 3, \textit{QNNVerifier} takes the annotated C code with the properties and performs invariant inference with Frama-C. An example of the result, with intervals for each neuron, is given in Figure~\ref{fig:insert_c_assumes}. These intervals are then automatically inserted back in the C code, encoded as \texttt{assume} instructions (see Figure \ref{fig:insert_c_invariant}).

\begin{figure}[ht]
\begin{lstlisting}[numbers=left,xleftmargin=2em,frame=single,framexleftmargin=1.5em, basicstyle=\tiny]  
float x0 = Frama_C_float_interval( 0 , 60760);
float x1 = Frama_C_float_interval( -3.141592 , 3.141592);
float x2 = Frama_C_float_interval( -3.141592 , 3.141592);
float x3 = Frama_C_float_interval( 100 , 1200);
float x4 = Frama_C_float_interval( 0 , 1200 );
\end{lstlisting}
\caption{Example of intervals provided by Frama-C. \label{fig:insert_c_assumes}}
\end{figure}

In step 4, the fixed-point operations specified by the user are replaced with their operational model, and the AFs are replaced by look-up tables. We provide pre-computed tables for the popular \textit{sigmoid} and \textit{tanh} AFs to ensure $\epsilon \leq 0.002$, but we leave the user the option of adding their look-up tables for custom AFs.

In step 5, \textit{QNNVerifier} invokes ESBMC to verify the annotated C file, which contains quantized ANN implementation and safety properties. Users can run a batch of verification processes in parallel. If ESBMC cannot find any counterexamples, the ANN is safe and ready to be deployed. Otherwise, a counterexample is provided, and the user must re-train the network or change its representation to make it safe.
Our tool includes examples of pre-annotated C files: all of our experiment files are available (see Section \ref{sec:exp}), with their corresponding 8-, 16-, and 32-bit fixed-point versions.

%-------------------------------------------------------
\section{Evaluation and Benchmarks}
\label{sec:exp}
%-------------------------------------------------------

\paragraph{Benchmarks.} We evaluate \textit{QNNVerifier} on three robustness benchmarks, which are representative of small to medium ANNs. The first is an ANN with $3$ layers and \textit{tanh} AFs trained on the Iris dataset. The second is an image classification ANN with $4$ layers and \textit{sigmoid} AFs trained on a character recognition dataset. The third is an ANN with $6$ layers of $300$ neurons each and ReLU AFs from the standard AcasXu benchmark. We conduct our experiments on an Intel(R) Xeon(R) CPU E$5$-$2620$ v$4$ @ $2$.$10$GHz with $128$ GB of RAM and Linux CentOS 7 (Core). See~\cite{sena2021verifying} for more information.

%\textbf{\textit{SMT solvers.}}

\paragraph{SMT solvers.} The verification performance of \textit{QNNVerifier} depends on the SMT solvers in the backend. We compare four of them: Bitwuzla, Boolector, Yices, and Z3. We show that Bitwuzla and Boolector have almost identical execution time, Yices is considerably faster across our whole verification suite, and Z3 struggles to complete most of the runs~\cite{sena2021verifying}.

%\textbf{\textit{Quantization.}}

\paragraph{Quantization.} Recall that verifying QNNs is PSPACE-hard in the worst-case scenario~\cite{Henzinger2020}, but empirical studies have shown a positive correlation between the word length and the total verification time~\cite{Giacobbe2020}. Using \textit{QNNVerifier}, we find that this correlation holds only for short word lengths (less than $16$ bits) and specific safety properties. Similarly, our results do not indicate relevant variations in safety for QNN with different quantization formats~\cite{sena2021verifying}.

%\textbf{\textit{Comparison with SOTA.}}

\paragraph{Comparison with SOTA} \textit{QNNVerifier}, despite targeting the ANN implementation, is not slower than state-of-the-art tools that verify the ANN abstract mathematical model. For example, in our experiments with AcasXu~\cite{sena2021verifying}, \textit{QNNVerifier} is faster than the SMT tool Marabou \cite{Katz2019} and on par with the reachability-based tool Neurify~\cite{Wang2018}.

%-------------------------------------------------------
\section{Conclusions}
\label{sec:conc}
%-------------------------------------------------------

\textit{QNNVerifier} is the first open-source verification tool capable of handling the finite word-length effects of quantized ANNs (fixed- and floating-point). Furthermore, \textit{QNNVerifier} is the only tool that supports all non-linear activation functions, and gives the user maximum freedom in specifying bespoke safety properties. Additionally, \textit{QNNVerifier} supports the largest number of ANN formats, including ONNX, NNET, Keras, Tensorflow, or even native C code. All of these unique functionalities ensure wide applicability of the tool.

\textit{QNNVerifier} verifies the source code of quantized ANNs with state-of-the-art software verification techniques, including interval analysis, fixed-point operational models and SMT-based model checking. Thanks to this, \textit{QNNVerifier} can efficiently verify ANNs with different quantization levels and non-linear activation functions. Moreover, \textit{QNNVerifier} is faster than existing SMT-based verification tools and on par with reachability-based tools, even though these other tools do not consider quantization effects.

%\textit{QNNVerifier} exploits state-of-the-art program analysis techniques to verify the safety of QNNs, including interval analysis, incremental solving, and expression simplification. \textit{QNNVerifier} is the verification tool that supports most formats of ANN, including ONNX, NNET, Keras, Tensorflow, or even native C code. Additionally and a unique feature, input ANNs can include any non-linear activation function. Furthermore, users can specify any safety property in plain C code, which can be checked using IEEE 754 floating-point and any fixed-point precision. Our results indicate that \textit{QNNVerifier} can efficiently verify ANNs with different quantization levels and non-linear AFs. Finally, \textit{QNNVerifier} is faster than other SMT-based verification tools and on par with reachability-based tools, even though it is the only tool to support quantization.

%To sum up, \textit{QNNVerifier} has significant unique functionality. First, it is the first open-source verification tool capable of handling the finite word-length effects of quantized ANNs. Second, \textit{QNNVerifier} is one of the few tools that supports all non-linear activation functions. Third, \textit{QNNVerifier} is the only tool that gives the user maximum freedom in specifying bespoke safety properties. Finally, \textit{QNNVerifier} supports the largest number of ANN formats to ensure wide applicability.
%-------------------------------------------------------
%\section*{Acknowledgments}
%\label{sec:ack}
%-------------------------------------------------------

% Bibliography
\bibliographystyle{ACM-Reference-Format}
\bibliography{references}

\end{document}